%% file: egpaper_final.tex
\documentclass[10pt,twocolumn,letterpaper]{article}
\usepackage[pdftex]{graphicx}

\usepackage{iccv}
\usepackage{times}
\usepackage{epsfig}
\usepackage{graphicx}
\usepackage{amsmath}
\usepackage{amssymb}
\usepackage{pifont}
\newcommand{\xmark}{\textcolor{red}{\ding{55}}}%
\newcommand{\cmark}{\textcolor{green}{\ding{51}}}%

\usepackage{colortbl}
\usepackage{multicol}
\usepackage{multirow}
\usepackage{xcolor}
\newcommand{\pub}[1]{\color{gray}{\tiny{[{#1}]}}}

\usepackage[pagebackref=true,breaklinks=true,letterpaper=true,colorlinks,bookmarks=false]{hyperref}

\iccvfinalcopy 

\def\iccvPaperID{****} 

\ificcvfinal\fi

\begin{document}

\title{TransHuman: A Transformer-based Human Representation for \\ Generalizable Neural Human Rendering}

\author{Xiao Pan$^{1,2,*}$, Zongxin Yang$^1$, Jianxin Ma$^2$, Chang Zhou$^2$, Yi Yang$^{1,\dag}$\\
$^{1}$ReLER Lab, CCAI, Zhejiang University, China \\
$^{2}$Alibaba DAMO Academy, China \\
{\tt\small \{xiaopan, yangzongxin, yangyics\}@zju.edu.cn  \{majx13fromthu, ericzhou.zc\}@alibaba-inc.com}
}

\maketitle
\ificcvfinal\thispagestyle{empty}\fi

\renewcommand{\thefootnote}{*}
\footnotetext{Work done during an internship with Alibaba DAMO Academy.}

\renewcommand{\thefootnote}{\dag}
\footnotetext{Corresponding author.}

\begin{abstract}
\input{./Sections/0_Abstract}
\end{abstract}


\input{./Sections/1_Introduction}
\input{./Sections/2_Relatedwork.tex}
\input{./Sections/3_Method.tex}
\input{./Sections/4_Experiment.tex}
\input{./Sections/5_Conclusion}

{\small
\bibliographystyle{ieee_fullname}
\bibliography{ref}
}

\clearpage
\appendix

\input{./Sections/6_Appendix}

\end{document}

%% file: Sections/0_Abstract.tex
   In this paper, we focus on the task of generalizable neural human rendering which trains conditional Neural Radiance Fields (NeRF) from multi-view videos of different characters. 
   To handle the dynamic human motion, previous methods have primarily used a SparseConvNet (SPC)-based human representation to process the painted SMPL. 
   However, such SPC-based representation i) optimizes under the volatile observation space  which leads to the pose-misalignment between training and inference stages, and ii) lacks the global relationships among human parts that is critical for handling the incomplete painted SMPL. 
   Tackling these issues, we present a brand-new framework named TransHuman, which learns the painted SMPL under the canonical space and captures the global relationships between human parts with transformers. Specifically, TransHuman is mainly composed of Transformer-based Human Encoding (TransHE), Deformable Partial Radiance Fields (DPaRF), and Fine-grained Detail Integration (FDI). TransHE first processes the painted SMPL under the canonical space via transformers for capturing the global relationships between human parts. Then, 
   DPaRF binds each output token with a deformable radiance field for encoding the query point under the observation space. Finally, the FDI is employed to further integrate fine-grained information from reference images.  Extensive experiments on ZJU-MoCap and H36M show that our TransHuman achieves a significantly new state-of-the-art performance with high efficiency. Project page: \url{https://pansanity666.github.io/TransHuman/}
    \vspace{-4mm}

%% file: Sections/1_Introduction.tex
\section{Introduction}
Rendering free-viewpoint videos of dynamic human performers in high fidelity is vital for many applications such as  mixed reality, gaming, and telepresence. Recent works~\cite{peng2021neural,peng2021animatable,weng2022humannerf,te2022NeuralCapture} integrate the Neural Radiance Fields (NeRF)~\cite{mildenhall2021nerf} technology with parametric human prior models (\eg, SMPL~\cite{loper2015smpl}) for handling the dynamic human body and achieve fair novel view synthesis results. However, the tedious per-subject optimization and the requirement of dense training views largely hinder the application of such methods. Targeting these issues and inspired by the recent success of generalizable NeRF~\cite{yu2021pixelnerf,mvsnerf,wang2021ibrnet} on static scenes, the task of generalizable neural human rendering is proposed~\cite{kwon2021nhp}, which trains conditional NeRF across multi-view human videos, and can generalize to a new subject in a single feed-forward manner given sparse reference views as input. 

\input{./Figures/idea_illustration}

Previous methods for generalizable neural human rendering~\cite{chen2022gpnerf,kwon2021nhp} mainly employ the SparseConvNet (SPC)~\cite{liu2015sparse}-based  human representation (upper row of Fig. ~\ref{fig_idea}) which first project deep features from reference images onto the vertices of fitted SMPL and then diffuse them to nearby regions via SPC. The final representation is achieved via the trilinear sampling  in the discrete 3D feature volume. Such SPC-based representation mainly suffers from the following two aspects: (i) \textit{Volatile observation learning.} The SPC-based one optimizes under the observation space that contains varying poses. This leads to the pose misalignment during training and inference stages, and therefore limits the generalization ability. 
(ii) \textit{Limited local receptive fields.} 
As shown in Fig.~\ref{fig_idea}, due to the heavy self-occlusion of dynamic human bodies, the painted SMPL templates are usually incomplete. While, as a 3D convolution network, the limited local receptive fields of SPC make it sensitive to the incomplete input, especially when the occluded regions are large. 

To address the aforementioned issues, we propose to first process the painted SMPL with transformers under the \textit{static canonical space} to remove the pose misalignment between training and inference stages and capture the \textit{global relationships} between human parts. Then, a deformation from the canonical to the observation space is required to fetch the human representation of a query point (sampling points on rays) under the observation space. Finally, the fine-grained information directly achieved from the observation space should be further included to the coarse human representation to complement the details. 


Motivated by this, we present the TransHuman, a brand-new framework that shows superior generalization ability with high efficiency. 
TransHuman is mainly composed of Transformer-based Human Encoding (TransHE), Deformable Partial Radiance Fields (DPaRF), and Fine-grained Detail Integration (FDI). (i) \textit{TransHE.} TransHE is a pipeline that processes the painted SMPL under the canonical space with transformers~\cite{dosovitskiy2020ViT}. The core of this pipeline includes a canonical body grouping strategy for the avoidance of semantic ambiguity, and a canonical learning scheme to ease the learning of global relationships. 
(ii) \textit{DPaRF.} DPaRF deforms the output tokens of TransHE from the canonical space to the observation space and gets a robust human representation for a query point from marched rays. As shown in Fig. \ref{fig_idea}, the main idea is to bind each token (representing a certain human part) with a radiance field whose partial coordinate system deforms as the pose changes, and the query point is encoded via the coordinates under the deformed partial coordinate systems.
(iii) \textit{FDI.} With TransHE and DPaRF, the human representation contains coarse information with human priors yet limited fine-grained details directly captured from the observation space. Therefore, similar to ~\cite{kwon2021nhp}, we propose to further integrate the detailed information from the pixel-aligned features at the guidance of the human representation.

Extensive experiments on ZJU-MoCap~\cite{peng2021neural} and H36M~\cite{ionescu2013human36m} demonstrate the superior generalization ability and high efficiency of TransHuman which attains a new state-of-the-art performance and outperforms previous methods by significant margins, \eg, $+2.20$ PSNR and $-45\%$ LPIPS on ZJU-MoCap~\cite{peng2021neural}  under the pose generalization setting. 



Our contributions are summarized as follows:
\begin{itemize}
    \item We propose a brand-new framework TransHuman for the challenging generalizable neural human rendering task which attains a significantly new state-of-the-art performance with high efficiency.

    \item We propose to process the painted SMPL under the canonical space to remove the pose misalignment during training and inference  stages and deform it back to the observation space via DPaRF for robust query point encoding. 
    
    \item To the best of our knowledge, we make the first attempt to explore the transformers technology around the painted SMPL for capturing the global relationships between human parts.



    
\end{itemize}


\input{./Figures/overview}

%% file: Figures/idea_illustration.tex
\begin{figure}[t]
\small
\centering
\includegraphics[width=\linewidth]{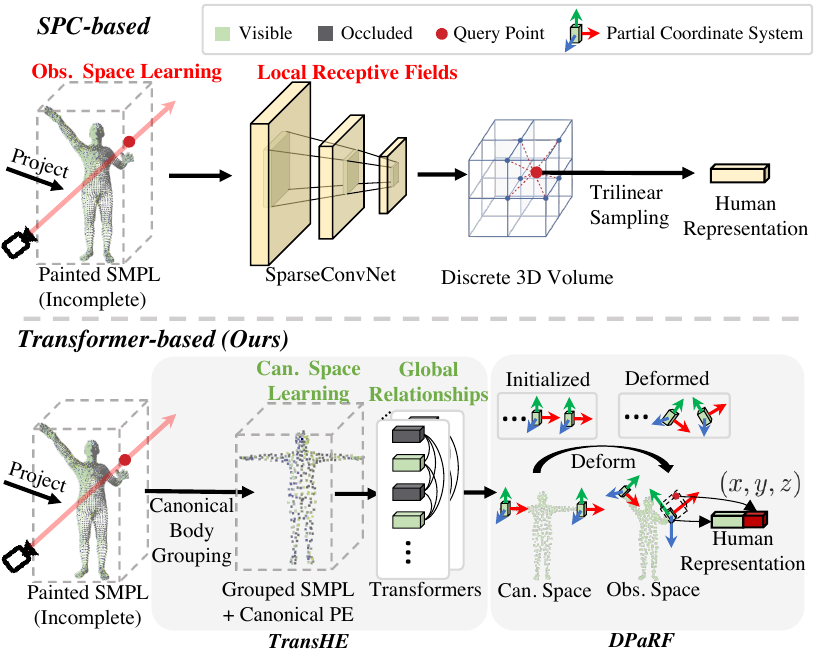}
\caption{\textbf{Comparisons between existing SPC-based and our transformer-based human representations.} Given the incomplete painted SMPL, the SPC-based one optimizes under the varying observation space with limited receptive fields from 3D convolution. Instead, our transformer-based one optimizes under the canonical space with global relationships between human parts.
}
\label{fig_idea}
\vspace{-4mm}
\end{figure}

%% file: Figures/overview.tex
\begin{figure*}[t]
\small
\centering
\includegraphics[width=1.0\linewidth]{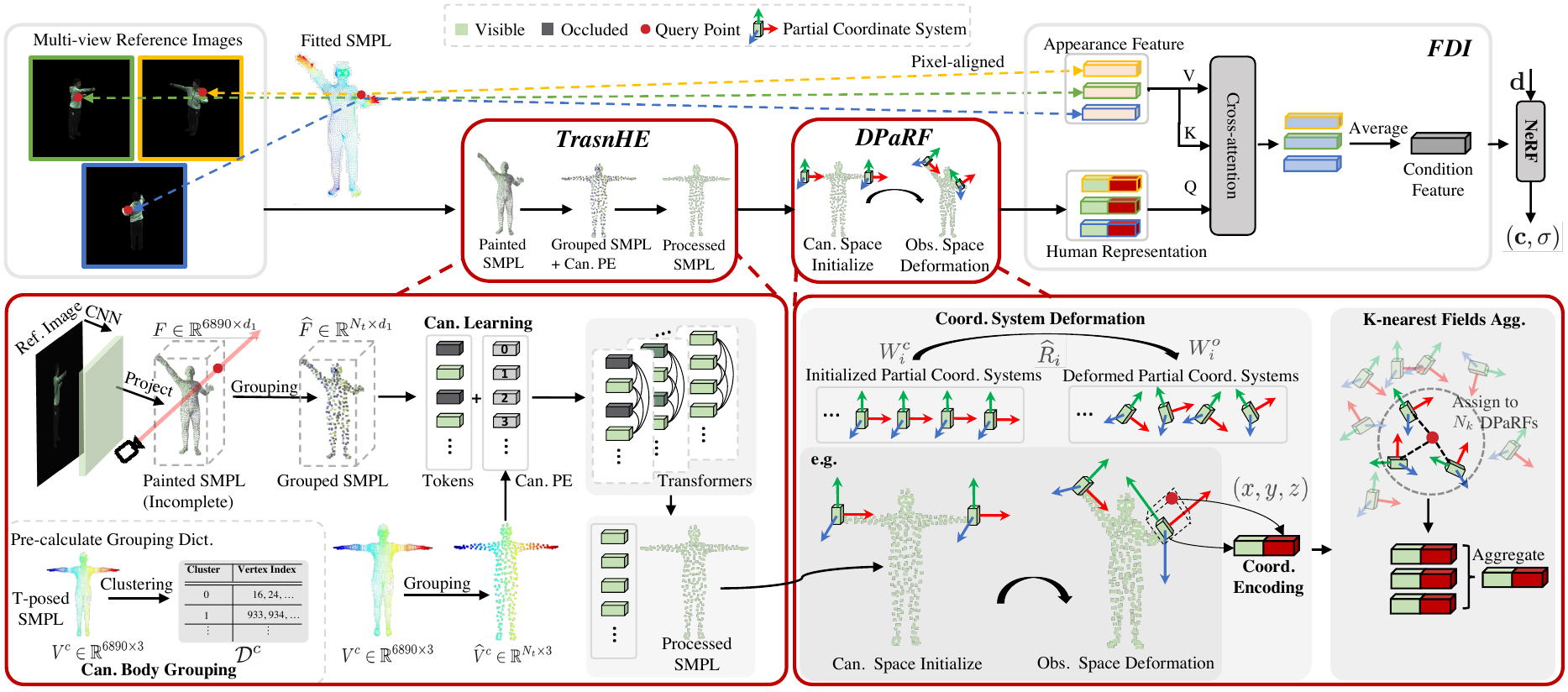}
\caption{\textbf{Overview of TransHuman.} TransHE first builds a pipeline for capturing the global relationships between human parts via transformers under the canonical space. Then, DPaRF deforms the coordinate system from the canonical back to the observation space and encodes a query point as an aggregation of coordinates and condition features.
 Finally, FDI further gathers the fine-grained information of the observation space from the pixel-aligned appearance feature under the guidance of human representation. }
\label{fig_overview}
 \vspace{-4mm}
\end{figure*}

%% file: Sections/2_Relatedwork.tex
\section{Related Work}

\subsection{Human Performance Capture}
\vspace{-2mm}
Synthesizing novel views for human performer is a long-standing topic in computer vision and graphics.  Traditional methods~\cite{HumanRender_convention_dou2016fusion4d,HumanRender_convention_collet2015high,HumanRender_convention_depthsensors_guo2019relightables,HumanRender_convention_depthsensors_debevec2000acquiring} typically require expensive hardware like depth sensors for getting reasonable results. With the recent success of Neural Radiance Fields (NeRF)~\cite{mildenhall2021nerf,barron2021mip}, many works~\cite{peng2021neural,peng2021animatable,weng2022humannerf,te2022NeuralCapture} have attempted to learn the 3D human representation from image inputs via differentiable rendering.
However, they require tedious per-subject optimization on dense training images, and can not generalize to unseen subjects, which largely confines the real-world applications. 

To tackle this issue and inspired by the recent advances of generalizable NeRF methods~\cite{yu2021pixelnerf,mvsnerf,wang2021ibrnet}, the generalizable neural human rendering task is explored~\cite{kwon2021nhp,MPSNERF,chen2022gpnerf,zhao2022humannerf}, 
At the core of this task is to properly exploit the human prior from the pre-fitted parametric human model. 
One line of works~\cite{zhao2022humannerf,MPSNERF}  take the parametric human model as the medium of the deformation between observation and canonical spaces using blend skinning technology~\cite{LBS_huang2020arch,LBS_lewis2000pose,LBS_liu2021neural}, and optimize conditional NeRF under a canonical pose. 
Instead, another line of works~\cite{kwon2021nhp,chen2022gpnerf} directly diffuse the painted parametric human model under the observation space via SparseConveNet (SPC)~\cite{liu2015sparse} for a human representation with approximate priors, and the final condition feature for a query point is the hybrid of human representation and pixel-aligned features. Obviously, a high-quality human representation is critical in this paradigm, yet the SPC-based one optimizes under the varying observation space, lacks the global perspective, and is restricted by the trilinear sampling in discrete 3D volumes. 

Targeting these issues, we present TransHuman with an advanced human representation based on transformers~\cite{vaswani2017attention,touvron2021DeiT,dosovitskiy2020ViT}, and outperforms the previous state-of-the-art methods by significant margins.

\subsection{Transformers with Neural Radiance Fields}
\vspace{-2mm}
With the significant advances of the transformer architecture~\cite{devlin2018bert,dosovitskiy2020ViT,caron2021DINO,radford2021CLIP},
several works~\cite{lin2023visionnerf,johari2022geonerf,reizenstein2021common,wang2021ibrnet,jain2021putting,xu2022sinnerf} have attempted to introduce it with NeRF technology. Specifically,  \cite{lin2023visionnerf} combines transformers with CNN~\cite{he2016Resnet} as a stronger feature extractor for reference images, \cite{johari2022geonerf,reizenstein2021common,wang2021ibrnet} use transformers as the aggregator of source view features, and 
\cite{jain2021putting,xu2022sinnerf} introduce the pre-trained transformers~\cite{radford2021CLIP,caron2021DINO} as a semantic prior to relieve the dense requirement of training views.

Differently, in this paper, we make the first attempt to apply the transformer technology around the surface of painted SMPL for a stronger human representation that captures the global relationship between human parts. 


%% file: Sections/3_Method.tex
\section{Method}

\noindent \textbf{Overview.} The task of generalizable neural human rendering targets on learning conditional NeRF across multi-view videos of different subjects, which can generalize to unseen subjects in a single feed-forward pass given sparse reference views. At the core of the task is to get a high-quality condition feature that contains accurate subject information for each query point sampled on rays. To this end, we propose a novel framework named TransHuman which shows superior generalization ability. As shown in Fig.~\ref{fig_overview}, TransHuman is mainly composed of three aspects: Transformer-based Human Encoding (TransHE), Deformable Partial Radiance Fields (DPaRF), and Fine-grained Detail Integration (FDI). \S~\ref{Sec_TransHE} introduces the TransHE which builds a pipeline for capturing the global relationships between human parts via transformers under the canonical space. \S~\ref{Sec_DPaRF} demonstrates the DPaRF which deforms the processed SMPL back to the observation space and fetch a robust human representation. \S~\ref{Sec_FDI} presents the FDI module that further gathers the fine-grained information directly from the observation space with the guidance of human representation. After that, we introduce the volume rendering  in \S~\ref{Sec_rendering}, and the training and inference pipelines in \S~\ref{sec_losses}.

\subsection{Transformer-based Human Encoding}
\vspace{-2mm}
\label{Sec_TransHE}
For simplicity, we start by introducing the process of a single reference image that is applicable for all other views, and the multi-view aggregation will be detailed in \S~\ref{Sec_FDI}.
Given a reference images $I$ for a certain time step and its corresponding pre-fitted SMPL model $V^o \in \mathbb{R}^{6890\times3}$ under the observation pose~\footnote{We use the SMPL coordinate system unless otherwise specified.}, we first project the $d_1$-dimensional deep features of $I$ extracted by CNN to the vertices of $V^o$ based on the camera information, and get the painted SMPL $F \in \mathbb{R}^{6890\times d_1}$. 
Previous methods~\cite{kwon2021nhp,chen2022gpnerf} have mainly employed the SPC~\cite{liu2015sparse} to diffuse the painted SMPL to nearby space (Fig.~\ref{fig_idea}). However, they optimize under the varying observation space which leads to the pose misalignment between training and inference stages, and the limited receptive fields of 3D convolution blocks make it sensitive to the incomplete painted SMPL input caused by the heavy self-occlusions of human bodies. Tackling these issues, we present a pipeline named Transformer-based Human Encoding (TransHE) that captures the global relationships between human parts under the canonical space. The key of TranHE includes a canonical body grouping strategy for avoiding the semantic ambiguity and a canonical learning scheme to ease the optimization and improve the generalization ability.  

\noindent \textbf{Canonical Body Grouping.}  Directly taking all the vertex features of $F$ as input tokens of transformers is neither effective considering the misalignment between fitted SMPL and the ground truth body, nor efficient due to the large vertex number, \ie, $6890$. A possible solution is to directly perform the grid voxelization~\cite{mao2021votr} on $F$ under the observation pose. However, due to the complex human poses, this will lead to the semantic ambiguity issue. 
More concretely, the gathered vertices in each voxel are highly different as the pose changes (\ie, temporal semantic variance), and a voxel might include vertices from dispersed semantic parts (\ie, spatial semantic entanglement), as illustrated in Fig. \ref{fig_semantic_ambiguity}.

To tackle this issue, we propose that grouping the vertices under the canonical space and then applying this canonical grouping to all the observation poses is a better choice. Compared with the varying observation poses, the canonical pose is both \textit{static} and more \textit{stretched}, which can largely relieve the semantic ambiguity issue via the consistent split among different poses (\ie, temporal semantic consistency) and more disentangled semantics in each voxel (\ie, spatial semantic disentanglement), as shown by the right part of Fig.~\ref{fig_semantic_ambiguity}. 

 \input{./Figures/Semantic_Ambiguity}

Formally, we first process the canonically posed (T-posed) SMPL  $V^{c} \in \mathbb{R}^{6890\times3} $ with a clustering algorithm (\eg, k-means~\cite{ahmed2020k}) based on the 3D coordinates, and get a grouping dictionary $\mathcal{D}^{c}$ caching the indexes of the SMPL vertices that belong to the same cluster, as illustrated in Fig.~\ref{fig_overview}. Notice that we only need to calculate $\mathcal{D}^{c}$ once before training. 
Then, for each iteration, the features from the same cluster are aggregated via average pooling: 
\begin{equation}
    \widehat{F} = \mathcal{G}_{\mathcal{D}^{c}}(F), \;\;\; \widehat{F} \in \mathbb{R}^{N_t \times d_1},
\end{equation}
where $N_t$ is the number of clusters (tokens), and $\mathcal{G}_{\mathcal{D}^{c}}(\cdot)$ indicates indexing based on $\mathcal{D}^{c}$ and then performing average pooling in each cluster. 


\noindent \textbf{Canonical Learning.} 
After grouping, we now have a decent number of input tokens, and the next question is about the choice of position embedding for each token. 
Since we need the condition feature of a query point under the observation space, a possible choice is to directly learn under the observation space (same as SPC-based methods~\cite{kwon2021nhp,chen2022gpnerf}) and use the 3D coordinates of each token under the observation pose as the position information, \ie, $\widehat{V}^o = \mathcal{G}_{\mathcal{D}^{c}}(V^o) \in \mathbb{R}^{N_t \times 3}$. However, except for the pose misalignment issue mentioned previously, 
$\widehat{V}^o$ is also varying for different time steps, which leads to the unfixed patterns of position embeddings that make it harder to capture the global relationships between human parts. 

Hence, to address these issues, we propose to learn the global relationships under the static canonical space for removing the pose-misalignment and easing the learning of global relationships: 
\begin{equation}
    \widehat{F}^{'} = \mathcal{T}(\widehat{F}, \gamma_1(\widehat{V}^c)),
\end{equation}
where $\widehat{V}^c = \mathcal{G}_{\mathcal{D}^{c}}(V^c)$ is the token positions under the canonical space, $\gamma_1(\cdot): \mathbb{R}^{3 \rightarrow d_1}$ represents the positional encoding used in the original NeRF~\cite{mildenhall2021nerf}, $\mathcal{T}(\cdot): \mathbb{R}^{d_1 \rightarrow d_1}$ indicates the transformers, and $\widehat{F}^{'} \in \mathbb{R}^{N_t \times d_1}$ is the output tokens with learned global relationships between each other.  



\subsection{Deformable Partial Radiance Fields}
\vspace{-2mm}
\label{Sec_DPaRF}
For deforming the processed SMPL back to the observation space and get a robust human representation, we present the Deformable Partial Radiance Fields (DPaRF). The main idea of DPaRF is to bind each output token of TransHE with a conditional partial radiance field for a certain semantic part whose coordinate system deforms as the pose changes under the observation space, and the query points from rays are encoded as the coordinates under the deformed coordinate system, as shown in Fig.~\ref{fig_overview}. 

\noindent \textbf{Coordinate System Deformation.} 
Given the $i$-th token $\widehat{F}^{'}_{i} \in \mathbb{R}^{d_1}$ from the TransHE output, a coordinate system $W^c_i \in \mathbb{R}^{3 \times 3 }$ is initialized under the canonical space which takes $\widehat{V}^c_{i} \in \mathbb{R}^{3}$ as the origin~\footnote{Without loss of generality, we set  $W_i$ as the identity matrix for all the tokens for simplicity.}. Then, as the pose changes under the observation space, we rotate  $W^c_i$ with the rotation matrix $\widehat{R}_i \in \mathbb{R}^{3 \times 3 } $ of token $i$:
\begin{equation}
    W_i^o = \widehat{R}_i W_i^c,
\end{equation}
where $\widehat{R}_i$ is the  averaged rotation matrix for vertices belonging to the $i$-th token, \ie, $ \widehat{R} = \mathcal{G}_{\mathcal{D}^c}(R) \in \mathbb{R}^{N_t \times 3 \times 3 } $, and $R \in \mathbb{R}^{6890 \times 3 \times 3 }$ can be calculated via blending the rotation matrices of $24$ joints with the blend weights provided by SMPL~\cite{loper2015smpl}. 


\noindent\textbf{Coordinate Encoding.}
After that, for a query point $\mathbf{p}$ sampled from the rays under the observation space, we get its coordinate   $\overline{\mathbf{p}}_i$ under the DPaRF of the $i$-th token with:
    \begin{equation}
          \overline{\mathbf{p}}_i =  W_i^o (\mathbf{p} - \widehat{V}_i^o).
    \end{equation}
And the final fetched human representation from the DPaRF of the $i$-th token is:
\begin{equation}
\label{eq_humanrepresentation}
    \mathbf{h}_i = [\widehat{F}^{'}_{i}; \gamma_2(\overline{\mathbf{p}}_i)],   \;\;\; \mathbf{h}_i \in \mathbb{R}^{d_2}, 
\end{equation}
where $[;]$ indicates the concatenation, and $\widehat{F}^{'}_{i}$ is the condition feature for the $i$-th DPaRF. 


    \noindent \textbf{K-nearest Fields Aggregation.}
Finally, for a more robust representation, we assign a query point $\mathbf{p}$  to its $N_k$ nearest DPaRFs, and aggregate them based on the distances:
\begin{equation}
\mathbf{h} = \sum_{i=1}^{N_k} softmax(- \frac{\| \mathbf{p} - \widehat{V}_i^o \| _2}{ \sum_i \| \mathbf{p} - \widehat{V}_i^o\|_2 }) \mathbf{h}_i,
\;\;\; \mathbf{h} \in \mathbb{R}^{d_2}.
\end{equation}

\subsection{Fine-grained Detail Integration}
\label{Sec_FDI}
\vspace{-2mm}

With TransHE and DPaRF, for a query point $\mathbf{p}$, we can actually achieve a set of human representations from $N_v$ reference views $ \mathbf{h}^{1:N_v} = \{ \mathbf{h}^j \}_{j=1}^{N_v} \in \mathbb{R}^{N_v \times d_2} $ 
 following the same procedure.  $\mathbf{h}^{1:N_v}$ contains coarse information with human priors (\eg,  geometry constraints and certain color information) yet lacks the fine-grained information (\eg, lighting, textures) for high-fidelity novel view synthesis. Therefore, inspired by~\cite{kwon2021nhp}, we further integrate the fine-grained information from the pixel-aligned appearance feature $\mathbf{a}^{1:N_v} = \{ \mathbf{a}^j\}_{j=1}^{N_v} \in \mathbb{R}^{Nv \times d_2} $ at the guidance of human representation $\mathbf{h}^{1:N_v}$. 

 \noindent\textbf{Fine-grained Appearance Features.}
 For the appearance features, instead of directly using projected deep features from CNN, \ie, the one used when painting SMPL, we additionally concatenate the projected  RGB-level information from the raw images and then fuse them with a fully connected layer $FC(\cdot): \mathbb{R}^{3+d_1 \rightarrow d_2}$. 
 The projected RGB features can complement the misaligned and lost details caused by the down-sample operation in CNN. 

 \noindent\textbf{Coarse-to-fine Integration.}
Then, we employ a cross-attention module which takes the human representation $\mathbf{h}^{1:N_v}$ as the query, and the appearance feature $\mathbf{a}^{1:N_v}$ as the key and value, and get the integrated feature $\mathbf{f}^{1:N_v} \in \mathbb{R}^{N_v \times d_2}$. The final condition feature  $\mathbf{f} \in \mathbb{R}^{d_2}$ of query point $\mathbf{p}$ is achieved via the average pooling on the view dimension: $\mathbf{f} = \sum_{j=1}^{N_c} \frac{1}{N_c}\mathbf{f}^{j}$. 



\subsection{Volume Rendering}
 \label{Sec_rendering}
 \vspace{-2mm}
 \noindent \textbf{Desnity \& Color Prediction.}
 The final density $\sigma(\mathbf{p}) \in \mathbb{R}^1$ and color $\mathbf{c}(\mathbf{p}) \in \mathbb{R}^3$ are predicted as: 
\begin{equation}
\begin{aligned}
    & \sigma(\mathbf{p}) = MLP_{\sigma}(\mathbf{f}), \\
    & \mathbf{c}(\mathbf{p}) = MPL_{\mathbf{c}}(\mathbf{f},\gamma_3( \mathbf{d})),
\end{aligned}
\end{equation}
where $MLP_{\sigma}$ and $ MLP_{\mathbf{c}}$ are NeRF MLPs for density and color predictions, respectively, and $\mathbf{d}$ is the unit view direction of the ray. 

\noindent \textbf{Differentiable Rendering.}
 Then, for a marched ray $\textbf{r}(z) = \textbf{o} + z \textbf{d}$, where $\textbf{o} \in  \mathbb{R}^3$ represents the camera center, and $z \in  \mathbb{R}^1$ is the depth between a pre-defined bounds $[z_n, z_f]$, its color $\textbf{C}(\textbf{r})$ is calculated via the differentiable volume rendering~\cite{mildenhall2021nerf}:
\begin{equation}
    \label{accumulate}
    \textbf{C}(\textbf{r}) = \int_{z_n}^{z_f} T(z) \sigma (z) \textbf{c}(z) dz,
\end{equation}
where $T(z) = exp (- \int_{z_n}^z \sigma(s) ds )$ represents the probability that the ray travels from $z$ to $z_n$. 

\subsection{Training \& Inference}
\label{sec_losses}
\vspace{-1mm}
\noindent\textbf{Training Losses.} We compare the rendered pixel colors with the ground truth ones for supervision. Similar to \cite{weng2022humannerf}, we employ the MSE loss for pixel-wise and perceptual loss~\cite{zhang2018unreasonable} for patch-wise supervision, which is more robust to misalignments. The random patch sampling~\cite{weng2022humannerf} is employed for supporting perceptual loss training.  The overall loss is:
\begin{equation}
    \mathcal{L} = \mathcal{L}_{MSE} + \lambda \mathcal{L}_{PER},
\end{equation}
where we set $\lambda=0.1$ by default.

\noindent\textbf{Inference.} During the inference stage, for each time step, $N_v$ reference views are provided and the rendered target views are compared with the ground truth ones for calculating the metrics. 
Notably, GP-NeRF~\cite{chen2022gpnerf} has proposed a fast rendering scheme that leverages the coarse geometry prior from the 3D feature volume to filter out useless points. Similarly, our framework also supports such strategy by simply using the SMPL template as the geometry prior instead (detailed in the appendix). 


%% file: Figures/Semantic_Ambiguity.tex
\begin{figure}[t]
\small
\centering
\includegraphics[width=0.95\linewidth]{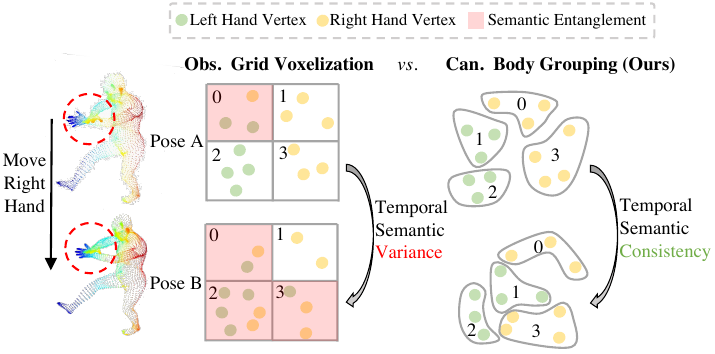}
\caption{\textbf{2D illustration of the semantic ambiguity issue.} Naive grid voxelization under the observation space leads to spatial semantic entanglement and temporal semantic variance issues, while the semantics with our canonical body grouping strategy is temporally consistent and spatially disentangled. }
\label{fig_semantic_ambiguity}
 \vspace{-4mm}
\end{figure}

%% file: Sections/4_Experiment.tex
\section{Experimental Results}
\input{./Tables/ZJU_MoCap_SOTA_V1}

\subsection{Experimental Settings}
\vspace{-2mm}
\noindent \textbf{Datasets.} We benchmark on ZJU-MoCap~\cite{peng2021neural} and H36M~\cite{ionescu2013human36m} for verifying the effectiveness of our TransHuman. 

(i) ZJU-MoCap~\cite{peng2021neural} provides multi-view videos of $10$ human subjects with $23$ synchronized  cameras, together with the pre-fitted SMPL parameters and human masks. Each video spans between $1000$ to $2000$ frames and contains complicated motions like ``Taichi" and ``Twirl".  
Following~\cite{kwon2021nhp,chen2022gpnerf}, $10$ subjects are split into $7$ source subjects (ZJU-$7$) and $3$ target subjects (ZJU-$3$), and each subject is further divided into training and testing parts.
We strictly follow the officially released human split from \cite{kwon2021nhp} for training and testing. We refer the detailed split information to the appendix. To prove that our method can welly handle the incomplete painted SMPL, we additionally report the performance of the one-shot generalization setting, \ie, only 1 reference view is provided during inference. 

(ii) H36M~\cite{ionescu2013human36m} records multi-view videos with 4 cameras and includes multiple subjects with complex motions. We use the preprocessed one by~\cite{peng2021animatable} which contains representative subjects S1, S5, S6, S7, S8, S9, S11, and their corresponding SMPL parameters and human masks. We verify the cross-dataset generalization ability with H36M, \ie, trained on ZJU-MoCap and then directly inference on H36M. The first 3 views are taken as the reference views, and the last one is used as the target view.



\noindent \textbf{Evaluation Metrics.}
For novel view synthesis, we report the commonly used Peak Signal-to-Noise Ratio (PSNR), Structural Similarity Index Measure (SSIM)~\cite{PSNR_wang2004image}, and Learned Perceptual Image Patch Similarity (LPIPS)~\cite{zhang2018unreasonable} as the evaluation metrics. For 3D reconstruction, following~\cite{kwon2021nhp,chen2022gpnerf}, we only report the qualitative results since ground truth meshes are unavailable.

\subsection{Implementation Details}
\vspace{-2mm}
In line with ~\cite{kwon2021nhp},  we take the ResNet-18~\cite{he2016Resnet} (only the first $3$ layers are used) as the CNN for extracting the deep features from reference images and set the multi-view number $N_v = 3$. The number of clusters (tokens) in human body grouping is set as $N_t = 300$, and the light-weight ViT-Tiny~\cite{dosovitskiy2020ViT} is employed as the transformer module. Each query point is assigned with $N_k = 7$ DPaRFs. Following ~\cite{kwon2021nhp,chen2022gpnerf}, we train on ZJU-MoCap with $512\times512$ 
resolutions, and for each ray we sample $64$ points by default during both the training and inference stages.

 \input{./Figures/ZJU_SOTA}
\input{./Figures/3D_rec}

\subsection{Comparisons with State-of-the-art}
\vspace{-2mm}
\noindent \textbf{Baselines.}
Following ~\cite{kwon2021nhp,chen2022gpnerf}, we compare with both per-subject optimization methods~\cite{peng2021neural,neural_textures,wu2020NHR,lombardi2019neuralvolume} and generalizable methods~\cite{raj2021pva,yu2021pixelnerf,keypointNeRF,kwon2021nhp,chen2022gpnerf}. For per-subject optimization methods, an individual model is trained on the training part of each subject. Notably, previous state-of-the-art methods for generalizable neural human rendering~\cite{kwon2021nhp,chen2022gpnerf} actually use different human splits in their officially released code and are not in line with the one used in their papers (performance is not reproducible). Hence, for fair comparisons, we  \textbf{unify them under the released human split of NHP~\cite{kwon2021nhp}}. Specifically, we report the performance of NHP~\cite{kwon2021nhp} using the official checkpoint, and re-run the official code of GP-NeRF~\cite{chen2022gpnerf} under the unified human split. Note that, GP-NeRF has employed an overfitting trick which we think is unreasonable, \ie, they overfit the test reference views  instead of randomly sampling during the training stage. This trick leaks the test information to the training stage, therefore we remove it in our re-running. We also provide the comparisons under the released human split of GP-NeRF with the overfitting trick, where our method outperforms it consistently by large margins. 

\noindent \textbf{Novel View Synthesis.}
We compare the quantitative results with previous state-of-the-art methods in Table~\ref{tab:ZJU_SOTA}. Obviously, we outperform them by significant margins under all the settings. 
Notably, for the identity generalization setting, the per-subject methods are directly trained on the target subjects while our method is only trained on the source subjects, 
yet we still outperform them by large margins, \ie, $+3.27$ in PSNR. Compared with the recent SPC-based generalizable methods~\cite{kwon2021nhp,chen2022gpnerf}, our method also shows healthy margins, \ie, $+2.20$ PSNR and $-45\%$ LPIPS compared with the second-best under the pose generalization setting. 
For the more challenging cross-dataset generalization setting, we also outperform the baseline methods remarkably albeit these two datasets~\cite{peng2021neural,ionescu2013human36m} have significantly different distributions, which proves the superior generalization ability of our TransHuman.

 The qualitative comparisons are illustrated in Fig. \ref{fig_zju_sota}, where our TransHuman gives significantly better details and body geometry. 
 We attribute this to the careful design of our framework, \ie, the global human representation brings more complete body geometry, the canonical learning scheme gives better generalization ability, and FDI further includes more fine-grained details like textures and lighting.
 

\noindent \textbf{3D Reconstruction.}
The 3D reconstruction results are illustrated in Fig.~\ref{fig_3d}. Compared with NHP~\cite{kwon2021nhp} that uses the SPC-based human representation, our method achieves a more complete and fine-grained geometry with details like wrinkles. 

\subsection{Ablation Studies}
\vspace{-2mm}
Following~\cite{kwon2021nhp}, we perform ablation studies under the identity generalization setting. Due to the limited space, we refer more detailed ablation studies to the appendix. 

\noindent \textbf{Ablation of TransHE.}
We first study the effectiveness of canonical body grouping and canonical learning scheme in Table~\ref{tab:abl_transhe}. When performing the body grouping under the observation space with grid voxelization (``obs. body grouping"), the performance suffers a significant drop from $26.15$ to $25.28$ in PSNR. As introduced in \S~\ref{Sec_TransHE}, performing grouping under the observation space leads to the semantic ambiguity issue, therefore leading to worse performance. Then,  ``obs. PE" changes the position embedding of input tokens from the canonical positions $\hat{V}^{c}$ to observation positions $\hat{V}^{o}$, and also observes a significant decrease, \eg, $-0.35$ in PSNR. The canonical learning scheme eases the optimization and removes the pose misalignment between training and inference stages, therefore leading to better performance. 

\noindent \textbf{Ablation of DPaRF.}
We verify the effectiveness of DPaRF in Table~\ref{tab:abl_dparf}. ``w/o coordinate" represents removing the coordinate part from the human representation.
As expected, the performance drops by significant margins ($-0.35$ in PSNR). Coordinates contain the accurate position information of query point in each DPaRF, therefore is important. ``absolute coordinate'' indicates using the absolute coordinate of query point, \ie, $\textbf{p}$ instead of $\overline{\textbf{p}}$ in Eq.~\ref{eq_humanrepresentation}, and the performance does not show significant improvement compared with ``w/o coordinate". This further proves the importance of using the coordinate under the deformed coordinate systems. Finally, ``w/o k-nearest fields" shows that the k-nearest fields aggregation design can bring certain improvement on all the metrics.

\input{./Tables/Ablation_TransHE}
\input{./Tables/Ablation_DPaRF}

\noindent  \textbf{Ablation of FDI.}
We first perform the ablation of FDI by individually removing the appearance feature part (``w/o $\textbf{a}$") or the human representation part (``w/o $\textbf{h}$"). As illustrated in Table~\ref{tab:abl_fdi}, merely using either of them gives an unsatisfactory performance. 
Then, ``w/o RGB" shows that  the raw RGB features can further bring a measure of improvement.

\input{./Tables/Ablation_FDI}
\input{./Tables/Ablation_SPC}

\noindent \textbf{Comparisons with SPC-based representation.}
To further verify the effectiveness of our proposed transformer-based human representation, we directly replace the TransHE and DPaRF modules with SPC and trilinear sampling in our code. We follow \cite{kwon2021nhp} to configure the SPC including the architecture and input resolution. As shown by Table \ref{tab:abl_spc}, our proposed transformer-based representation outperforms the SPC-based one by significant margins among all the metrics under a fair comparison setting.

\subsection{Efficiency Analysis}
\vspace{-1mm}

\input{./Tables/Efficiency}

We compare the efficiency of our method with previous state-of-the-art methods in Table~\ref{tab:efficiency} under the identity generalization setting (438 frames). For a fair comparison with the previously fastest method GP-NeRF~\cite{chen2022gpnerf} under the same inference time, we provide a fast version of our method by reducing the sampling points per ray from $64$ to $16$ during inference (``Ours-16pts"). Obviously, with the same inference time, our method still outperforms GP-NeRF by $0.84$ in PSNR albeit using merely $64\%$ parameters, $55\%$ inference memory, and $71\%$ training memory, and the performance can be further significantly improved with acceptable additional cost. This proves that our TransHuman is both effective and efficient.  
\vspace{-1mm}

%% file: Tables/ZJU_MoCap_SOTA_V1.tex
\begin{table*}[t]
\small
\centering
\setlength\tabcolsep{8pt}
\begin{tabular}{l|cc|c|cc|ccc} 
 \rowcolor[gray]{.9}
\hline
          & \multicolumn{2}{c|}{Dataset} & Per-subject &\multicolumn{2}{c|}{Unseen} &  \multicolumn{3}{c}{Results} \\  \rowcolor[gray]{.9} 
Method   & Train & Test &  training  & Pose & Subject & $\uparrow$ PSNR & $\uparrow$ SSIM & $\downarrow$ LPIPS \\ \hline
\hline
\multicolumn{9}{c}{\textit{Pose Generalization}} \\
NV~\pub{TOG19}~\cite{lombardi2019neuralvolume} &  ZJU-$7$&   ZJU-$7$ & \cmark   & \cmark & \xmark    & 22.00          & 0.818          & -   \\
NT~\pub{TOG19}~\cite{neural_textures} &  ZJU-$7$&   ZJU-$7$ & \cmark   & \cmark & \xmark    & 22.28  & 0.872  & -    \\
NHR~\pub{CVPR20}~\cite{wu2020NHR} &  ZJU-$7$&   ZJU-$7$ & \cmark   & \cmark & \xmark     & 22.31  & 0.871 & - \\
NB~\pub{CVPR21}~\cite{peng2021neural} &  ZJU-$7$&   ZJU-$7$ & \cmark  & \cmark & \xmark     & 23.79 &  0.887  & - \\ 
NHP~\pub{NIPS21}~\cite{kwon2021nhp} &  ZJU-$7$&   ZJU-$7$ & \xmark  & \cmark & \xmark     & 24.60 &  0.910 & 0.147  \\ 
GP-NeRF~\pub{ECCV22}~\cite{chen2022gpnerf} &  ZJU-$7$&   ZJU-$7$ & \xmark      & \cmark & \xmark    &      25.05 &  0.909 & 0.159 \\
\textbf{Ours} &  ZJU-$7$&   ZJU-$7$ & \xmark      & \cmark & \xmark    &      {\textbf{27.25}} &  {\textbf{0.936}} & \textbf{0.087 }\\
\hline \hline
\multicolumn{9}{c}{\textit{Identity Generalization}} \\
NV~\pub{TOG19}~\cite{lombardi2019neuralvolume} &  ZJU-$3$&   ZJU-$3$ & \cmark   & \cmark & \xmark  & 20.84   &  0.827  & -   \\
NT~\pub{TOG19}~\cite{neural_textures} &  ZJU-$3$&   ZJU-$3$ & \cmark   & \cmark & \xmark    & 21.92  & 0.873   & -    \\
NHR~\pub{CVPR20}~\cite{wu2020NHR} &  ZJU-$3$&  ZJU-$3$ & \cmark   & \cmark & \xmark    & 22.03  & 0.875 & -  \\
NB~\pub{CVPR21}~\cite{peng2021neural} &  ZJU-$3$&   ZJU-$3$ & \cmark  & \cmark & \xmark    & 22.88 &  0.880 & -  \\ 
PVA~\pub{arXiv21}~\cite{raj2021pva} &  ZJU-$7$&   ZJU-$3$ & \xmark  & \cmark & \cmark    & 23.15 &  0.866 & -  \\ 
PixelNeRF~\pub{CVPR21}~\cite{yu2021pixelnerf} &  ZJU-$7$&   ZJU-$3$ & \xmark  & \cmark & \cmark     & 23.17 &  0.869 & -  \\
KeyNeRF~\pub{ECCV22}~\cite{keypointNeRF} &  ZJU-$7$&   ZJU-$3$ & \xmark      & \cmark & \cmark  &  25.03 &  0.897 & - \\
GP-NeRF~\pub{ECCV22}~\cite{chen2022gpnerf} &  ZJU-$7$&   ZJU-$3$ & \xmark      & \cmark & \cmark  &  24.55 &  0.902 & 0.157 \\
NHP~\pub{NIPS21}~\cite{kwon2021nhp} &  ZJU-$7$&   ZJU-$3$ & \xmark  & \cmark & \cmark     & 24.94 &  0.905 & 0.144  \\
\textbf{Ours} &  ZJU-$7$&   ZJU-$3$ & \xmark  & \cmark & \cmark     & \textbf{26.15} &  \textbf{0.918} & \textbf{0.098}  \\ \hline

GP-NeRF\dag~\pub{ECCV22}~\cite{chen2022gpnerf} &  ZJU-$7$&   ZJU-$3$ & \xmark      & \cmark & \cmark  &  26.83  &  0.924 & 0.132 \\
\textbf{Ours\dag} &  ZJU-$7$&   ZJU-$3$ & \xmark  & \cmark & \cmark     & \textbf{27.55} &  \textbf{0.933} & \textbf{0.090}  \\

\hline \hline
\multicolumn{9}{c}{\textit{One-shot Generalization}} \\
NHP~\pub{NIPS21}~\cite{kwon2021nhp} &  ZJU-$7$ & ZJU-$3$ & \xmark  & \cmark & \cmark      & 23.20 &  0.877 & 0.182  \\
\textbf{Ours} &  ZJU-$7$ & ZJU-$3$ & \xmark      & \cmark & \cmark   &  \textbf{24.11} & \textbf{0.891} &  \textbf{0.142} \\ \hline \hline

\multicolumn{9}{c}{\textit{Cross-dataset Generalization}} \\
NHP~\pub{NIPS21}~\cite{kwon2021nhp} &  ZJU-$7$ & H36M & \xmark  & \cmark & \cmark      & 18.84 &  0.820 & 0.222  \\
\textbf{Ours} &  ZJU-$7$ & H36M & \xmark      & \cmark & \cmark   &  \textbf{20.48} & \textbf{0.856} &  \textbf{0.169} \\

\hline
\end{tabular}
\vspace{+0.5mm}
 \caption{
 \textbf{Comparisons of generalization ability with the state-of-the-art methods.} We achieve a significantly new sate-of-the-art performance compared with both generalizable~\cite{raj2021pva,yu2021pixelnerf,chen2022gpnerf,kwon2021nhp,keypointNeRF} and per-subject methods~\cite{lombardi2019neuralvolume,neural_textures,wu2020NHR,peng2021neural}. Following \cite{kwon2021nhp}, the per-subject optimization methods  are trained on the training part of each subject since they can not generalize to unseen subjects, which is actually an easier task. ``\dag" means using the officially released human split from GP-NeRF~\cite{chen2022gpnerf} and employing the overfitting trick used in GP-NeRF.  }
    \label{tab:ZJU_SOTA}
    \vspace{-4mm}
\end{table*}

%% file: Figures/ZJU_SOTA.tex
\begin{figure*}[t]
\small
\centering
\includegraphics[width=0.99\linewidth]{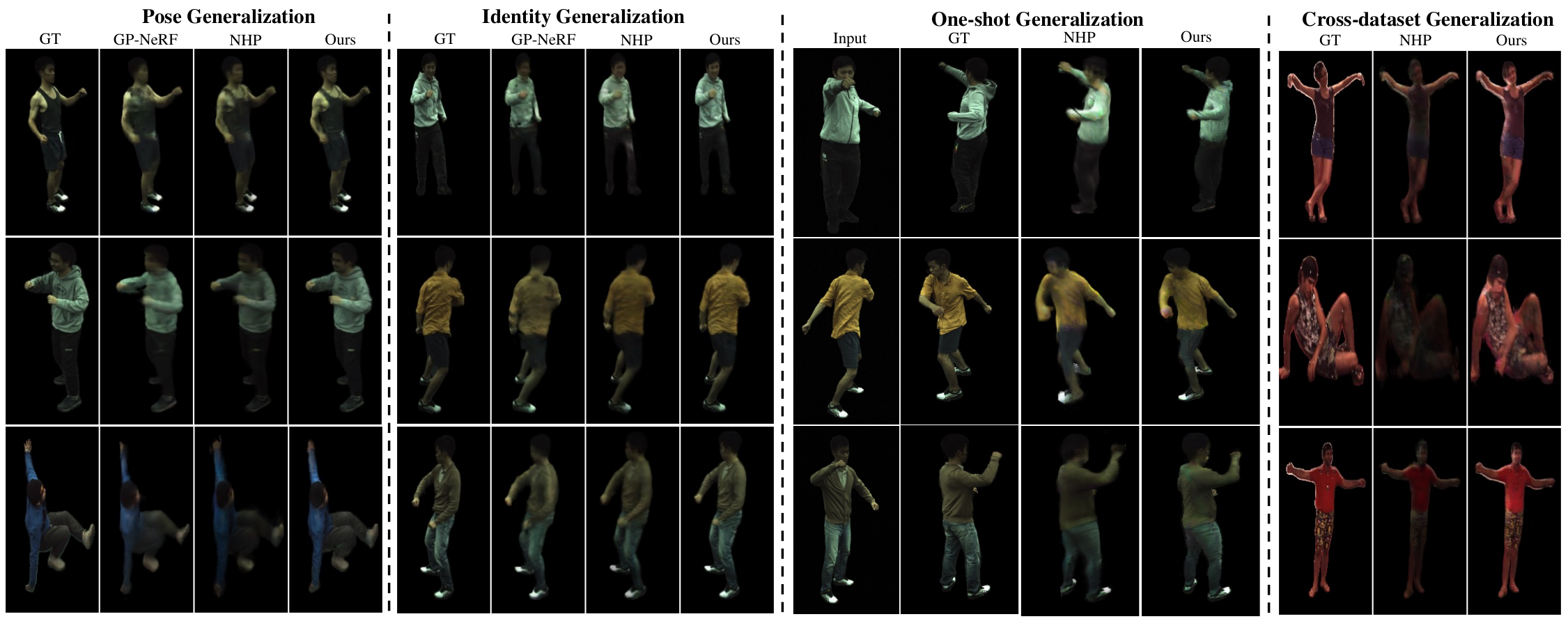}
\vspace{-1mm}
\caption{\textbf{Visualization comparisons with previous state-of-the-art methods on ZJU-MoCap (pose generalization, identity generalization) and H36M (cross-dataset generalization).} Our method shows significantly better generalization ability with better body geometry and more accurate details like textures and lighting.  }
\label{fig_zju_sota}
\vspace{-4mm}
\vspace{-1mm}
\end{figure*}

%% file: Figures/3D_rec.tex
\begin{figure}[t]
\small
\centering
\includegraphics[width=0.82\linewidth]{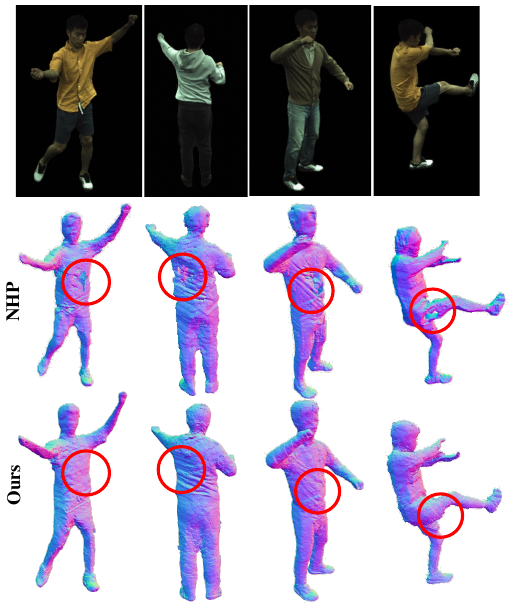}
\vspace{-2mm}
\caption{\textbf{3D reconstruction under the identity generalization setting.} Our method achieves more complete geometry with details like wrinkles compared with NHP~\cite{kwon2021nhp} which employs a SPC-based human representation.}
\label{fig_3d}
\vspace{-4mm}
\end{figure}

%% file: Tables/Ablation_TransHE.tex
\begin{table}[t]
\small
\setlength\tabcolsep{9.8pt}
\begin{tabular}{l|ccc}
\rowcolor[gray]{.9}
\hline 
Method &  $\uparrow$  PSNR   & $\uparrow$  SSIM  & $\downarrow$  LPIPS \\ \hline \hline 
obs. body grouping    & 25.28 & 0.909 & 0.111 \\
obs. PE   & 25.80 & 0.915 & 0.102\\ \hline

\begin{tabular}[c]{@{}l@{}}can. body grouping \\ + can. PE\end{tabular}   & \textbf{26.15} & \textbf{0.918} & \textbf{0.098} \\

\hline
\end{tabular}
 \caption{
     \textbf{Ablation of TransHE.} Our canonical body grouping together with the canonical learning scheme performs best. }
    \label{tab:abl_transhe}
    \vspace{-2mm}
\end{table}

%% file: Tables/Ablation_DPaRF.tex
\begin{table}[t]
\small
\setlength\tabcolsep{9.7pt}
\begin{tabular}{l|ccc}
\rowcolor[gray]{.9}
\hline 
Method &  $\uparrow$  PSNR   & $\uparrow$  SSIM  & $\downarrow$  LPIPS \\ \hline \hline 
w/o coordinate    & 25.80 & 0.912 & 0.123 \\
absolute coordinate   & 25.76 & 0.912 & 0.116\\
w/o k-nearest fields   & 26.05 & 0.916 & 0.099 \\ \hline
full model  & \textbf{26.15}
& \textbf{0.918} & \textbf{0.098} \\
\hline
\end{tabular}
 \caption{
     \textbf{Ablation of DPaRF.} Coordinate encoding is critical and the k-nearest fields aggregation can further bring improvements. }
    \label{tab:abl_dparf}
    \vspace{-2mm}
\end{table}

%% file: Tables/Ablation_FDI.tex
\begin{table}[t]
\small
\setlength\tabcolsep{14pt}
\begin{tabular}{l|ccc}
\rowcolor[gray]{.9}
\hline 
Method &  $\uparrow$  PSNR   & $\uparrow$  SSIM  & $\downarrow$  LPIPS \\ \hline \hline 
w/o $\mathbf{a}$    & 24.58 & 0.898 & 0.134  \\
w/o $\mathbf{h}$   & 24.66 & 0.897  & 0.143 \\ 
w/o RGB    & 26.05 & 0.917  & 0.101  \\ \hline
full model & \textbf{26.15} & \textbf{0.918} & \textbf{0.098} \\
\hline
\end{tabular}
 \caption{
     \textbf{Ablation of FDI.} Using the appearance feature and human representation individually leads to the drop of performance, and the raw RGB feature can bring certain improvement.  }
    \label{tab:abl_fdi}
    \vspace{-4mm}
\end{table}

%% file: Tables/Ablation_SPC.tex
\begin{table}[t]
\small
\setlength\tabcolsep{7pt}
\begin{tabular}{l|ccc}
\rowcolor[gray]{.9}
\hline 
Method &  $\uparrow$  PSNR   & $\uparrow$  SSIM  & $\downarrow$  LPIPS \\ \hline \hline 
SPC + trilinear & 25.14 & 0.907 & 0.102 \\ 
TransHE + DPaRF (ours) & \textbf{26.15} & \textbf{0.918} & \textbf{0.098} \\

\hline
\end{tabular}
 \caption{
     \textbf{Comparision with SPC-based representation.} Our transformer-based representation outperforms the SPC-based one significantly.  }
    \label{tab:abl_spc}
    \vspace{-2mm}
\end{table}

%% file: Tables/Efficiency.tex
\begin{table}[t]
\small
\setlength\tabcolsep{3.3pt}
\begin{tabular}{l|ccccc}
\rowcolor[gray]{.9}
\hline
Method     & Param. & \begin{tabular}[c]{@{}c@{}}Inference \\ Time\end{tabular} & \begin{tabular}[c]{@{}c@{}}Inference \\ Mem.\end{tabular}   & \begin{tabular}[c]{@{}c@{}}Training \\ Mem.\end{tabular} & PSNR  \\ \hline \hline
NHP~\cite{kwon2021nhp}        & \textbf{5.80M}  & 1h55min                                                   & 6.4GB                                                       & 12.2GB              & 24.94                                     \\
GP-NeRF~\cite{chen2022gpnerf}    & 9.52M  & \textbf{9min}                                                      & 10.3GB                                                      & 11.0GB                & 24.55                                   \\ 
\textbf{Ours-16pts} & 6.08M  & \textbf{9min}                                                      & \textbf{5.7GB}                                                      & \textbf{7.8GB}     & \textbf{25.39}    \\ \hline    \hline

Ours       & 6.08M  & 17min                                                     & 6.2GB                                                     & 7.8GB              & 26.15                                      \\ \hline
                                       
\end{tabular}
 \caption{
     \textbf{Efficiency comparisons under the identity generalization setting.} With the same inference time, our method outperforms GP-NeRF~\cite{chen2022gpnerf} significantly in PSNR albeit requiring fewer parameters and training/inference memory. The performance can further be greatly improved at the cost of certain additional inference time and minor inference memory. }
    \label{tab:efficiency}
    \vspace{-2mm}
\end{table}

%% file: Sections/5_Conclusion.tex
\section{Conclusion}
\vspace{-2mm}
In this paper, we propose a brand-new framework named TransHuman for the generalizable neural human rendering task. At the core of TransHuman is a canonically optimized human representation with global relationships between human parts captured by transformers which shows superior generalization ability compared with previous methods.  However, there are remaining challenges to be explored, such as the joint optimization of fitted SMPL and training on unconstrained multi-view capture setups. We hope that our efforts will motivate more researchers in the future.

%% file: Sections/6_Appendix.tex

\section{Progressive Rendering}
\input{./Tables/Appendix_Progressive}
\vspace{-2mm}
GP-NeRF~\cite{chen2022gpnerf} has proposed a progressive rendering strategy using the coarse geometry provided by the output 3D feature volume of SPC to reduce the number of rendering points. Although there is no SPC in our framework,  we find that simply using the fitted SMPL as the alternative works pretty well. Specifically, after sampling points on marched rays from the target view, we only render the points whose euclidean distance to the SMPL template is smaller than $0.1m$. Then, for these close points, we first get  the density values for all of them, and then only send part of them whose density value is larger than $0$ for the following color inference, which is in line with ~\cite{chen2022gpnerf}. The effectiveness of this strategy is illustrated in Table ~\ref{tab:progressive}. While without decreasing the performance, the inference time is reduced by around $70\%$. Notably, even without using such accelerating strategy, the inference is still over 2 times faster than NHP~\cite{kwon2021nhp} (56min \vs 1h55min, Table~\ref{tab:efficiency}). This strongly proves the efficiency of our method.

\section{Performance on Training Frames}
\input{./Tables/Appendix_Fitting}
\vspace{-2mm}
Following previous 
methods~\cite{kwon2021nhp,chen2022gpnerf}, we report the fitting performance on the training set in Table~\ref{tab:fitting}.  We achieve the best fitting performance among the generalizable methods, which shows the superior capacity of our method. 

\section{Additional Ablation Studies}
\vspace{-2mm}
We provide more detailed ablation studies in this section. 

\subsection{Influence of Cluster Number $N_t$}
\input{./Tables/Appendix_Cluster_Number}
\vspace{-2mm}
We study the influence of cluster (token) number by varying it sequentially as $\{100, 300, 500, 1000\}$. As shown in Table~\ref{tab:cluster_number}, too large cluster number does not bring further improvement. As mentioned in \S~\ref{Sec_TransHE}, there exists misalignment between the fitted SMPL and the ground truth body. Larger cluster number may also include more misleading information, and we only intend to take the human representation as the coarse-level guidance, therefore we set $N_t=300$.

\subsection{Influence of K-nearest Number $N_k$}
\input{./Tables/Appendix_K_Number}
\vspace{-2mm}
We show the influence of k-nearest number $N_k$ in Table~\ref{tab:k_number}. When using no k-nearest fields aggregation, \ie, $N_k=1$, the performance suffers a relatively significant drop in PSNR. This shows that using k-nearest fields aggregation can improve the robustness of human representation. When $N_k>1$, the performance tends to be more stable, and we choose $N_k$ as $7$ since it gives the best performance. 

\subsection{Influence of Perceptual Loss}
\input{./Tables/Appendix_Perceptual_Loss}
\vspace{-2mm}
In Table~\ref{tab:percep}, we demonstrate the influence of perceptual loss. Obviously, perceptual loss can largely improve the LPIPS, \ie, make the results visually pleasing, while shows less effect on PSNR and SSIM. Without perceptual loss, we still outperform previous methods by consistent margins in PSRN and SSIM.

\subsection{Canonical K-means \vs Canonical Grid Voxelization }
\input{./Figures/kmeans_grid}
\input{./Tables/Appendix_kmeans_grid}
\vspace{-2mm}
In canonical body grouping, we employ the k-means clustering to get the grouping dictionary. Actually, using grid voxelization under the canonical space is also feasible. However, the uniform grid leads to the large variance of vertex number in each voxel considering the shape of human body, as shown in Fig.~\ref{fig:kmeans_grid}. Therefore, we use k-means instead for a more uniform split. As illustrated in Table~\ref{tab:kmeans_grid}, canonical k-means performs better than canonical grid voxelization.

\section{Additional Visualization Examples}
\vspace{-2mm}
We provide more visualization examples in this section. 

\subsection{Ablation of FDI}
\input{./Figures/Ablation_FDI}
\vspace{-2mm}
To better illustrate the functional difference between human representation $\textbf{h}$ and appearance feature $\textbf{a}$ in FDI, we provide the ablation in Fig.~\ref{fig_abl_fdi}.  Obviously, the human representation $\textbf{h}$ contains geometry constraints from human priors with coarse color information, while $\textbf{a}$ shows more vivid colors with poor geometry. Hence, we propose to take the coarse human representation as the guidance for integrating proper fine-grained details from the appearance feature.

\subsection{Comparisons with State-of-the-art}
\input{./Figures/Appendix_VIS_SOTA}
\vspace{-2mm}
We provide more comparison examples with previous state-of-the-art methods in Fig.~\ref{fig_appendix_vis_sota}. 

\section{Human Split}
\input{./Tables/Human_Split}
\vspace{-2mm}
 We list the detailed human split information in Table~\ref{tab:human_split}. We hope that it can serve as a standard split for the following researchers. The code will also be available upon acceptance.

%% file: Tables/Appendix_Progressive.tex
\begin{table}[t]
\small
\centering
\setlength\tabcolsep{6pt}
\begin{tabular}{l|cccc}
\rowcolor[gray]{.9}
\hline 
Method &  $\uparrow$  PSNR   & $\uparrow$  SSIM  & $\downarrow$  LPIPS & \begin{tabular}[c]{@{}c@{}}Inference \\ Time\end{tabular} \\ \hline \hline 
w/o progressive & \textbf{26.15} & \textbf{0.918} & \textbf{0.098}  & 56min \\
\textbf{full model}   & \textbf{26.15} & \textbf{0.918} & \textbf{0.098} & \textbf{17min}  \\
\hline 
\end{tabular}
 \caption{\textbf{Effectiveness of progressive rendering strategy.} The progressive rendering strategy can reduce the inference time by around $70\%$ while without influencing the performance.}
 \label{tab:progressive}
\end{table}

%% file: Tables/Appendix_Fitting.tex
\begin{table}[t]
\small
\centering
\setlength\tabcolsep{14.3pt}
\begin{tabular}{l|ccc}
\rowcolor[gray]{.9}
\hline 
Method &  $\uparrow$  PSNR   & $\uparrow$  SSIM  & $\downarrow$  LPIPS \\ \hline \hline 
NHP  & 25.65 &  0.917 & 0.148  \\ 
GP-NeRF & 26.46 &  0.918 & 0.158 \\
\textbf{Ours}     &      {\textbf{28.08}} &  {\textbf{0.939}} & \textbf{0.087}\\
\hline
\end{tabular}
 \caption{\textbf{Fitting performance on training frames.} Our method shows the best fitting ability compared with previous methods.}
    \label{tab:fitting}
    \vspace{-2mm}
\end{table}

%% file: Tables/Appendix_Cluster_Number.tex
\begin{table}[t]
\small
\centering
\setlength\tabcolsep{14pt}
\begin{tabular}{l|ccc}
\rowcolor[gray]{.9}
\hline 
Method &  $\uparrow$  PSNR   & $\uparrow$  SSIM  & $\downarrow$  LPIPS \\ \hline \hline 
$N_t$ = 100   & 26.04 &  0.917 & 0.100  \\ 
\textbf{$N_t$ = 300 } & \textbf{26.15} & \textbf{0.918} & \textbf{0.098} \\
 $N_t$ = 500  & 26.10 &  0.917 & 0.100  \\
  $N_t$ = 1000  & 26.07 &  0.917 & 0.100  \\
\hline 
\end{tabular}
 \caption{\textbf{Influence of cluster number $N_t$.} $N_t=300$ gives the best performance.}
 \label{tab:cluster_number}
\end{table}

%% file: Tables/Appendix_K_Number.tex
\begin{table}[t]
\small
\centering
\setlength\tabcolsep{15pt}
\begin{tabular}{l|ccc}
\rowcolor[gray]{.9}
\hline 
Method &  $\uparrow$  PSNR   & $\uparrow$  SSIM  & $\downarrow$  LPIPS \\ \hline \hline 
$N_k$ = 1   & 26.05 &  0.917 & 0.099  \\ 
$N_k$ = 3  & 26.11 &  0.918 & 0.100  \\
 $N_k$ = 5  & 26.13 &  0.918 & 0.100  \\
\textbf{$N_k$ = 7}   & \textbf{26.15} & \textbf{0.918} & \textbf{0.098}  \\
   $N_k$ = 9  & 26.10 &  0.917 & 0.100  \\
\hline 
\end{tabular}
 \caption{\textbf{Influence of k-nearest number $N_k$.} $N_k=7$ performs best. }
 \label{tab:k_number}
\end{table}

%% file: Tables/Appendix_Perceptual_Loss.tex
\begin{table}[t]
\small
\centering
\setlength\tabcolsep{13.2pt}
\begin{tabular}{l|ccc}
\rowcolor[gray]{.9}
\hline 
Method &  $\uparrow$  PSNR   & $\uparrow$  SSIM  & $\downarrow$  LPIPS \\ \hline \hline 
w/o $\mathcal{L}_{PER}$   &\textbf{26.16} &  0.916 & 0.146  \\ 
\textbf{full model} & {26.15} & \textbf{0.918} & \textbf{0.098} \\
\hline 
\end{tabular}
 \caption{\textbf{Influence of perceptual loss.} Perceptual loss mainly improves the LPIPS with less effect on PSNR and SSIM.  }
 \label{tab:percep}
\end{table}

%% file: Figures/kmeans_grid.tex
\begin{figure}[!t]
\centering
\small
\centering
\includegraphics[width=1.0\linewidth]{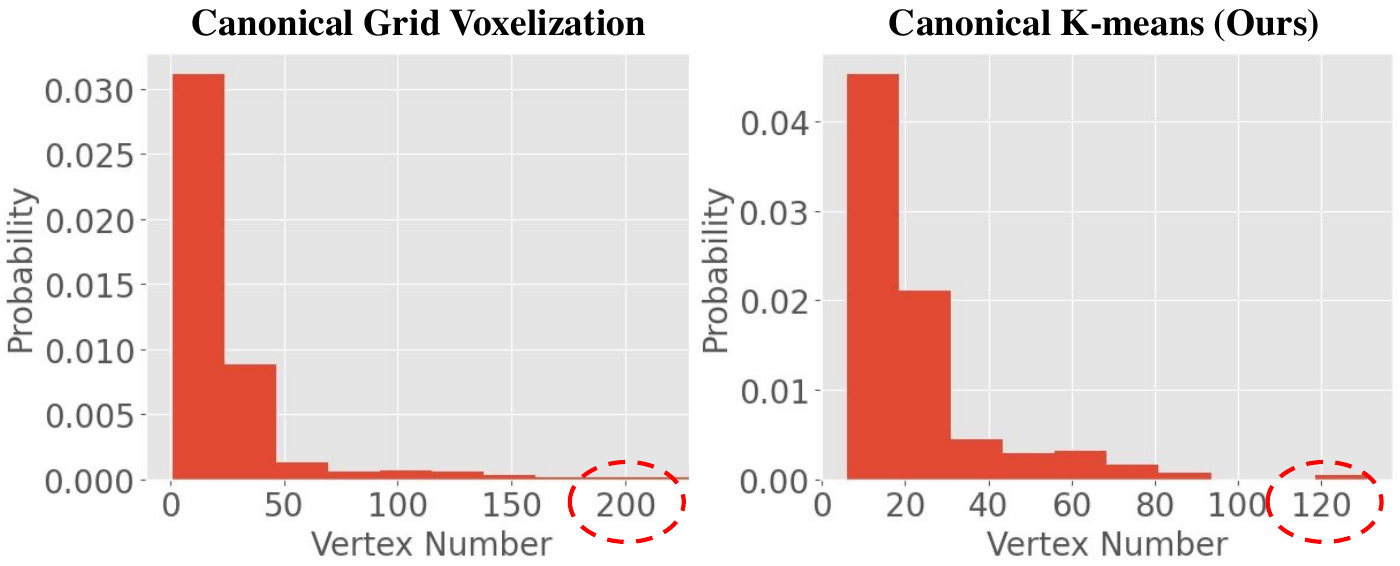}
\caption{\textbf{Comparisons of vertex number distributions between canonical grid voxelization and canonical k-means.} Canonical k-means gives more uniform split with smaller variance.}
\label{fig:kmeans_grid}
\end{figure}

%% file: Tables/Appendix_kmeans_grid.tex
\begin{table}[t]
\small
\centering
\setlength\tabcolsep{9pt}
\begin{tabular}{l|ccc}
\rowcolor[gray]{.9}
\hline 
Method &  $\uparrow$  PSNR   & $\uparrow$  SSIM  & $\downarrow$  LPIPS \\ \hline \hline 
can. grid voxelization & {26.01} & {0.917} & {0.100}   \\
\textbf{can. k-means (ours)}  & \textbf{26.15} & \textbf{0.918} & \textbf{0.098} \\
\hline 
\end{tabular}
\vspace{-1mm}
 \caption{\textbf{Comparisons between using k-means and grid voxelization in canonical body grouping.} }
 \label{tab:kmeans_grid}
\end{table}

%% file: Figures/Ablation_FDI.tex
\begin{figure}[t]
\small
\centering
\includegraphics[width=0.9\linewidth]{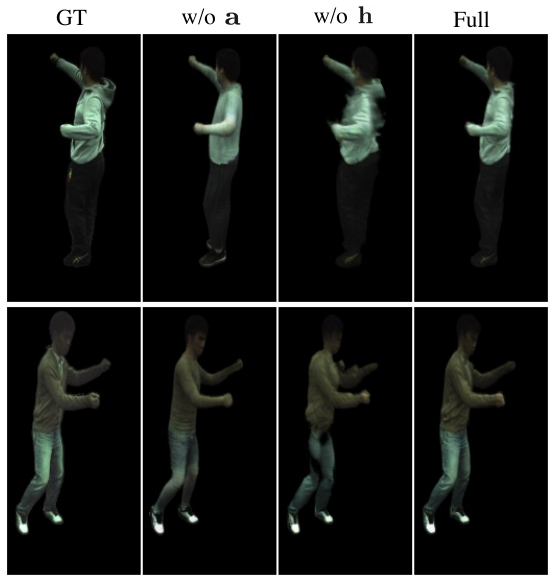}
\caption{ \textbf{Ablation of human representation $\textbf{h}$ and appearance feature $\textbf{a}$ in FDI.} Human representation $\textbf{h}$  provides geometry constraints from human priors and coarse color information, and further integrates fine-grained information from appearance features $\textbf{a}$ with FDI.}
\label{fig_abl_fdi}
\end{figure}

%% file: Figures/Appendix_VIS_SOTA.tex
\begin{figure*}[t]
\small
\centering
\includegraphics[width=1.0\linewidth]{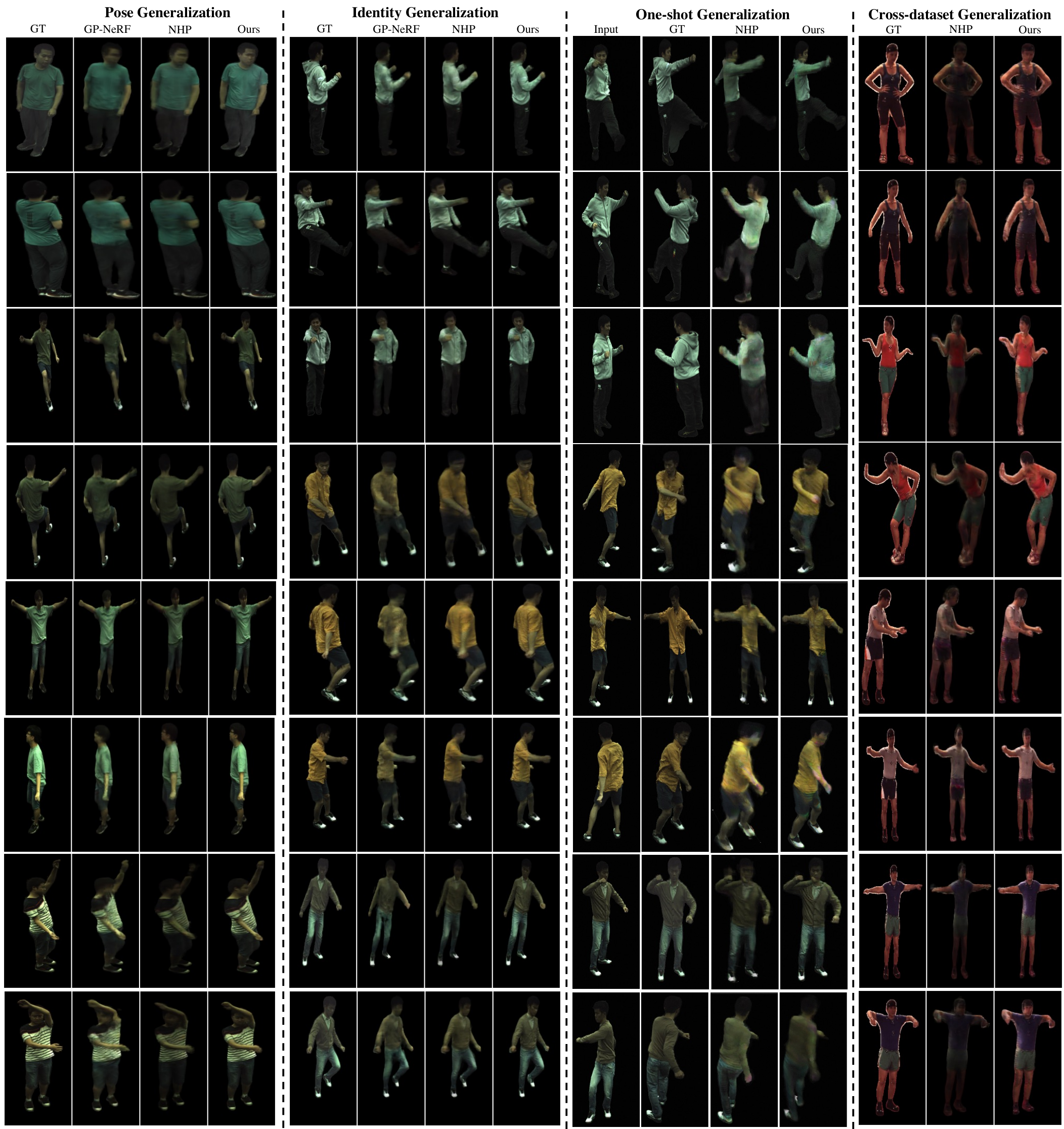}
\caption{ \textbf{Supplemented visualization examples on ZJU-MoCap~\cite{peng2021neural} (pose generalization, identity generalization, one-shot generalization) and H36M~\cite{ionescu2013human36m} (cross-dataset generalization).}}
\label{fig_appendix_vis_sota}
\end{figure*}

%% file: Tables/Human_Split.tex
\begin{table*}[t]
\small
\centering
\setlength\tabcolsep{8pt}
\begin{tabular}{c|lc|c|c|c|c}
\rowcolor[gray]{.9}
\hline
{Human ID} & {[Start, End)} & {Interval} & {Frame Number} & {Total Frames} & {Reference View} & {Target View} \\ \hline \hline
\multicolumn{7}{c}{\textit{Training Frames}}                                                                                                            \\ \hline
313               & [0, 60)               & 1                 & $60 \times 21$               & \multirow{7}{*}{7589} & Rand 3                  & Rand 1               \\
315               & [0, 400)              & 6                 & $67 \times 21$               &                       & Rand 3                  & Rand 1               \\
377               & [0, 300)              & 30                & $10 \times 23$               &                       & Rand 3                  & Rand 1               \\
386               & [0, 300)              & 6                 & $50 \times 23$               &                       & Rand 3                  & Rand 1               \\
390               & [700, 1000)           & 6                 & $50 \times 23$               &                       & Rand 3                  & Rand 1               \\
392               & [0, 300)              & 6                 & $50 \times 23$               &                       & Rand 3                  & Rand 1               \\
396               & [810, 1080)           & 5                 & $54 \times 23$               &                       & Rand 3                  & Rand 1               \\ \hline
\multicolumn{7}{c}{\textit{Pose Generalization}}                                                                                                        \\ \hline
313               & [60, 1060)            & 30                & $34 \times 6$                & \multirow{7}{*}{798}  & 0, 7, 15                & 3, 5, 10, 12, 18, 20 \\
315               & [400, 1400)           & 30                & $34 \times 6$                &                       & 0, 7, 15                & 3, 5, 10, 12, 18, 20 \\
377               & [300, 617)            & 30                & $11 \times 6$                &                       & 0, 7, 15                & 3, 5, 10, 12, 18, 20 \\
386               & [300, 646)            & 30                & $12 \times 6$                &                       & 0, 7, 15                & 3, 5, 10, 12, 18, 20 \\
390               & [0, 700)              & 30                & $24 \times 6$                &                       & 0, 7, 15                & 3, 5, 10, 12, 18, 20 \\
392               & [300, 556)            & 30                & $9 \times 6$                 &                       & 0, 7, 15                & 3, 5, 10, 12, 18, 20 \\
396               & [1080, 1350)          & 30                & $9 \times 6$                 &                       & 0, 7, 15                & 3, 5, 10, 12, 18, 20 \\ \hline
\multicolumn{7}{c}{\textit{Identity   Generalization}}                                                                                                  \\ \hline
387               & [0, 654)              & 30                & $22 \times 6$                & \multirow{3}{*}{438}  & 0, 7, 15                & 3, 5, 10, 12, 18, 20 \\
393               & [0, 658)              & 30                & $22 \times 6$                &                       & 0, 7, 15                & 3, 5, 10, 12, 18, 20 \\
394               & [0, 859)              & 30                & $29 \times 6$                &                       & 0, 7, 15                & 3, 5, 10, 12, 18, 20 \\ \hline
\multicolumn{7}{c}{\textit{One-shot   Generalization}}                                                                                                  \\ \hline
387               & [0, 654)              & 30                & $22 \times 6$                & \multirow{3}{*}{438}  & 0                       & 3, 5, 10, 12, 18, 20 \\
393               & [0, 658)              & 30                & $22 \times 6$                &                       & 0                       & 3, 5, 10, 12, 18, 20 \\
394               & [0, 859)              & 30                & $29 \times 6$                &                       & 0                       & 3, 5, 10, 12, 18, 20 \\ \hline 
\multicolumn{7}{c}{\textit{Cross-dataset   Generalization}}    \\ \hline
S1  & [0, 750)  & 150 & $5 \times 1$  & \multirow{7}{*}{54} & 0, 1, 2 & 3 \\
S5  & [0, 1250) & 150 & $9 \times 1$  &                     & 0, 1, 2 & 3 \\
S6  & [0, 750)  & 150 & $5 \times 1$  &                     & 0, 1, 2 & 3 \\
S7  & [0, 1500) & 150 & $10 \times 1$ &                     & 0, 1, 2 & 3 \\
S8  & [0, 1250) & 150 & $9 \times 1$  &                     & 0, 1, 2 & 3 \\
S9  & [0, 1300) & 150 & $9 \times 1$  &                     & 0, 1, 2 & 3 \\
S11 & [0, 1000) & 150 & $7 \times 1$  &                     & 0, 1, 2 & 3 \\ \hline

\end{tabular}
\vspace{+1mm}
 \caption{\textbf{Detailed human split.} {For ZJU-MoCap~\cite{peng2021neural} (training frames, pose, identity, and one-shot generalization), we follow the human split from the officially released code of NHP~\cite{kwon2021nhp}, while for H36M~\cite{ionescu2013human36m} (cross-dataset generalization), we follow the split from ~\cite{peng2021animatable}.} Note that in ZJU-MoCap, ``313" and ``315" only contains $21$ camera views, while the left ones have $23$ camera views.} 
\label{tab:human_split}
\end{table*}